%% file: main.tex
\documentclass{article} %
\usepackage{iclr2021_conference,times}

\input{math_commands.tex}

\def\onedot{.}
\def\eg{\emph{e.g}\onedot} 
\def\ie{\emph{i.e}\onedot} 
 
\def\etc{\emph{etc}\onedot}

\def\aka{\emph{a.k.a}\onedot}

\usepackage{hyperref}
\usepackage{url}
\usepackage{color}
\usepackage{algorithm}  
\usepackage{algorithmicx}
\usepackage{algpseudocode}
\usepackage{multirow}
\usepackage{graphicx}
\usepackage{booktabs}
\usepackage{placeins}
\usepackage{enumitem}
\usepackage{listings}
\usepackage{wrapfig}
\usepackage{pifont}
\usepackage{xcolor,colortbl}
\usepackage[symbol]{footmisc}

\definecolor{darkergreen}{RGB}{21, 152, 56}
\definecolor{red2}{RGB}{252, 54, 65}

\title{\vspace{-0.6cm}Dual-mode ASR: Unify and Improve Streaming ASR with Full-context Modeling}

\author{
\parbox{\linewidth}{\centering
Jiahui Yu\(^1\) \hspace{.7cm}
Wei Han\(^1\)\thanks{equal contribution} \hspace{.7cm}
Anmol Gulati\(^1{^\dagger}\) \hspace{.7cm}
Chung-Cheng Chiu\(^1\) \hspace{.7cm}
Bo Li\(^2\) \hspace{.7cm}\vspace{0.1cm}\\
Tara N.\ Sainath\(^2\) \hspace{.7cm}
Yonghui Wu\(^1\) \hspace{.7cm}
Ruoming Pang\(^1\) \hspace{.7cm}
}\\
\parbox{\linewidth}{\centering
\(^1\)Google Brain \hspace{.7cm} \(^2\)Google LLC\vspace{0.1cm}\\
\texttt{\{jiahuiyu, rpang\}@google.com}\vspace{-1cm}
}
}

\iclrfinalcopy %
\begin{document}

\maketitle

\begin{abstract}
\textit{Streaming} automatic speech recognition (ASR) aims to emit each hypothesized word as quickly and accurately as possible, while \textit{full-context} ASR waits for the completion of a full speech utterance before emitting completed hypotheses. In this work, we propose a unified framework, \textit{Dual-mode ASR}, to train a single end-to-end ASR model with shared weights for both streaming and full-context speech recognition. We show that the latency and accuracy of streaming ASR significantly benefit from \textit{weight sharing} and \textit{joint training} of full-context ASR, especially with \textit{inplace knowledge distillation} during the training. The Dual-mode ASR framework can be applied to recent state-of-the-art convolution-based and transformer-based ASR networks. We present extensive experiments with two state-of-the-art ASR networks, ContextNet and Conformer, on two datasets, a widely used public dataset LibriSpeech and a large-scale dataset MultiDomain. Experiments and ablation studies demonstrate that Dual-mode ASR not only simplifies the workflow of training and deploying streaming and full-context ASR models, but also significantly improves both emission latency and recognition accuracy of streaming ASR. With Dual-mode ASR, we achieve new state-of-the-art streaming ASR results on both LibriSpeech and MultiDomain in terms of accuracy and latency.
\end{abstract}

\input{sec1_intro}

\input{sec2_related_work}

\input{sec3_approach}

\input{sec4_experiments}

\input{sec5_conclusion}

\bibliography{iclr2021_conference}
\bibliographystyle{iclr2021_conference}

\end{document}

%% file: math_commands.tex
\usepackage{amsmath,amsfonts,bm}

\def\eqref#1{equation~\ref{#1}}

\def\1{\bm{1}}

\DeclareMathAlphabet{\mathsfit}{\encodingdefault}{\sfdefault}{m}{sl}
\SetMathAlphabet{\mathsfit}{bold}{\encodingdefault}{\sfdefault}{bx}{n}

%% file: sec1_intro.tex
\section{Introduction}
``Ok Google. Hey Siri. Hi Alexa.'' have featured a massive boom of smart speakers in recent years, unveiling a trend towards ubiquitous and ambient Artificial Intelligence (AI) for better daily lives. As the communication bridge between human and machine, low-latency streaming ASR (\aka, online ASR) is of central importance, whose goal is to emit each hypothesized word as quickly and accurately as possible on the fly as they are spoken. On the other hand, there are some scenarios where full-context ASR (\aka, offline ASR) is sufficient, for example, offline video captioning on video-sharing platforms. While low-latency streaming ASR is generally preferred in most of the speech recognition scenarios, it often has worse prediction accuracy as measured in Word Error Rate (WER), due to the lack of future context compared with full-context ASR. Improving both WER and emission latency has been shown to be highly challenging~\citep{he2019streaming, li2020towards, sainath2020streaming} in streaming ASR systems.

\begin{figure}[t]
\centering
\includegraphics[width=0.9\linewidth]{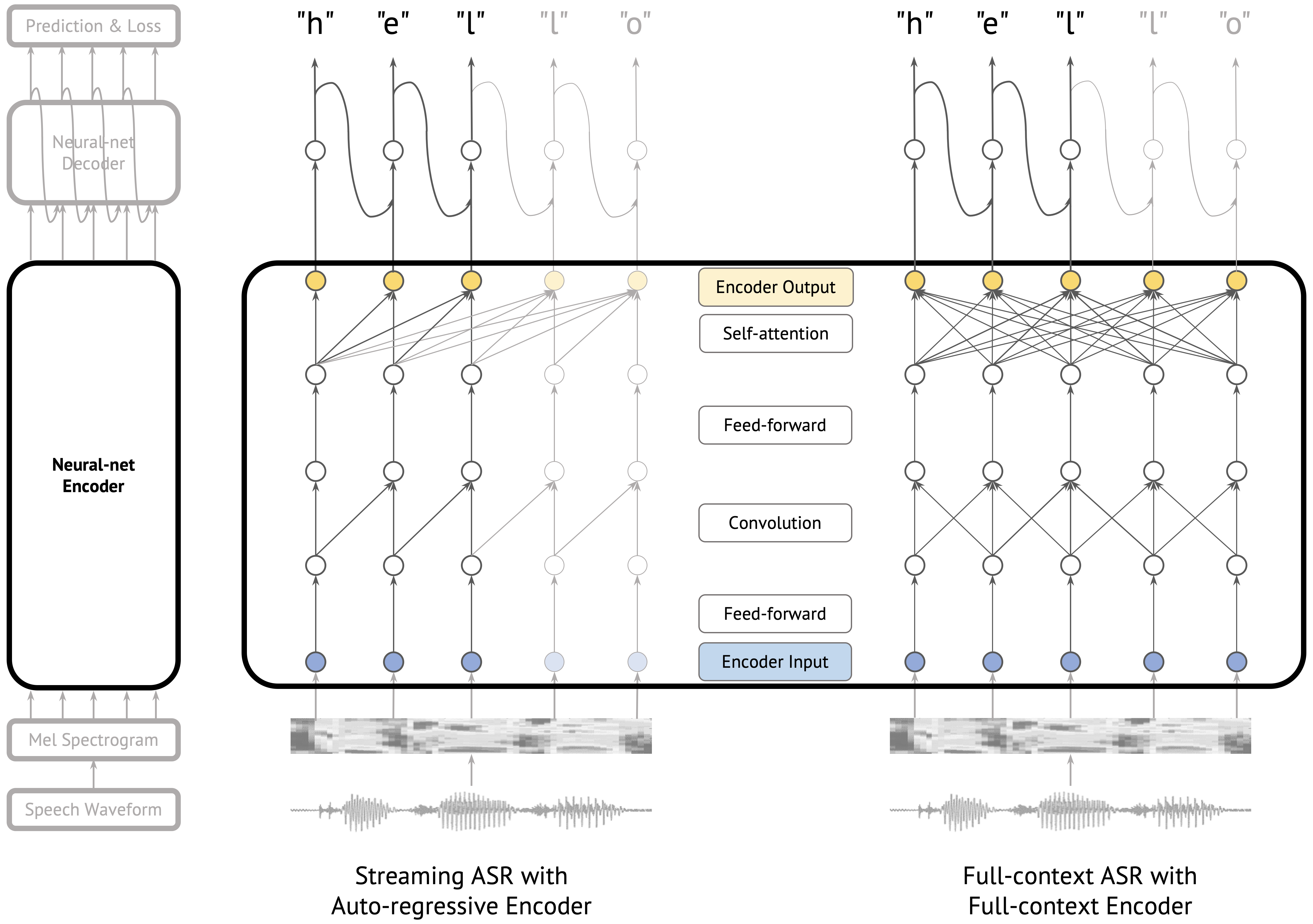}
\vspace{-0.3cm}
\caption{A simplified illustration of the similarity and difference between Streaming ASR and Full-context ASR networks. Modern end-to-end streaming and full-context ASR models share most of the neural architectures and training recipes in common, with the most significant difference in the \textbf{ASR encoder (highlighted)}. Streaming ASR encoders are auto-regressive models, with each prediction of the current timestep conditioned on previous ones (no future context). We show examples of feed-forward layer, convolution layer and self-attention layer in the encoder of streaming and full-context ASR respectively. With Dual-mode ASR, we unify them without parameters overhead.}
\vspace{-0.3cm}
\label{figs:overview_similarity}
\end{figure}

Since the acoustic, pronunciation, and language model (AM, PM, and LM) of a conventional ASR system have been evolved into a single end-to-end (E2E) all-neural network, modern streaming and full-context ASR models share most of the neural architectures and training recipes in common, such as, Mel-spectrogram inputs, data augmentations, neural network meta-architectures, training objectives, model regularization techniques and decoding methods. The most significant difference is that \textit{streaming ASR encoders are auto-regressive models}, with the prediction of the current timestep conditioned on previous ones (no future context is permitted). Specifically, let \(x\) and \(y\) be the input and output sequence, \(t\) as frame index, \(T\) as total length of frames. Streaming ASR encoders model the output \(y_t\) as a function of input \(x_{1:t}\) while full-context ASR encoders model the output \(y_t\) as a function of input \(x_{1:T}\). Streaming ASR encoders can be built with uni-directional LSTMs, causal convolution and left-context attention layers in streaming ASR encoders~\citep{chiu2018monotonic, fan2018online, han2020contextnet, gulati2020conformer, huang2020conv, moritz2020streaming, miao2020transformer, tsunoo2020streaming, zhang2020transformer, yeh2019transformer}. Recurrent Neural Network Transducers (RNN-T)~\citep{graves2012sequence} are commonly used as the decoder in both streaming and full-context models, which predicts the token of the current input frame based on all previous tokens using uni-directional recurrent layers. Figure~\ref{figs:overview_similarity} illustrates a simplified example of the similarity and difference between streaming and full-context ASR models with E2E neural networks.

Albeit the similarities, streaming and full-context ASR models are usually developed, trained, and deployed separately. In this work, we propose \textit{Dual-mode ASR}, a framework to unify streaming and full-context speech recognition networks with shared weights. Dual-mode ASR comes with many immediate benefits, including reduced model download and storage on devices and simplified development and deployment workflows. To accomplish this goal, we first introduce \textit{Dual-mode Encoders}, which can run in both streaming mode and full-context mode. Dual-mode encoders are designed to reuse the same set of model weights for both modes with zero or near-zero parameters overhead. We propose the design principles of a dual-mode encoder and show examples on how to design dual-mode convolution, dual-mode pooling, and dual-mode attention layers. We also investigate into different training algorithms for Dual-mode ASR, specifically, randomly sampled training and joint training. We show that joint training significantly outperforms randomly sampled training in terms of model quality and training stability. Moreover, motivated by \textit{Inplace Knowledge Distillation}~\citep{yu2019universally} in which a large model is used to supervise a small model, we propose to \textit{distill knowledge from the full-context mode (teacher) into the streaming mode (student) on the fly} during the training within the same Dual-mode ASR model, by encouraging consistency of the predicted token probabilities.

We demonstrate that the emission latency and prediction accuracy of streaming ASR significantly benefit from \textit{weight sharing} and \textit{joint training} of its full-context mode, especially with \textit{inplace knowledge distillation} during the training. We present extensive experiments with two state-of-the-art ASR networks, convolution-based ContextNet~\citep{han2020contextnet} and conv-transformer hybrid Conformer~\citep{gulati2020conformer}, on two datasets, a widely used public dataset LibriSpeech~\citep{panayotov2015librispeech} (970 hours of English reading speech) and a large-scale dataset MultiDomain~\citep{narayanan2018toward} (413,000 hours speech of a mixture across multiple domains including Voice Search, Farfield Speech, YouTube and Meetings). For each proposed technique, we also present ablation study and analysis to demonstrate and understand the effectiveness. With Dual-mode ASR, we achieve new state-of-the-art streaming ASR results on both LibriSpeech and MultiDomain in terms of accuracy and latency.

%% file: sec2_related_work.tex
\section{Related Work}

\textbf{Streaming ASR Networks.}
There has been a growing interest in building streaming ASR systems based on E2E Recurrent Neural Network Transducers (RNN-T)~\citep{graves2012sequence}. Compared with sequence-to-sequence models~\citep{chorowski2014end, chorowski2015attention, chorowski2016towards, bahdanau2016end, chan2016listen}, RNN-T models are naturally \textit{streamable} and have shown great potentials for low-latency streaming ASR~\citep{chang2019joint, he2019streaming, tsunoo2019towards, sainath2019two, shen2019lingvo, li2020towards, li2020parallel, sainath2020streaming, huang2020conv, moritz2020streaming, narayanan2020cascaded}. In this work, we mainly focus on RNN-T based models. \citeauthor{he2019streaming} specifically studied how to optimize the RNN-T streaming ASR model for mobile devices, and proposed a bag of techniques including using layer normalization and large batch size to stabilize training; using word-piece targets~\citep{wu2016google}; using a time-reduction layer to speed up training and inference; quantizing network parameters to reduce memory footprint and speed up computation; applying shallow-fusion to bias towards user-speciﬁc context. To support streaming modeling in E2E ASR models, various efforts have also been made by modifying attention-based models such as monotonic attention~\citep{colin17,chiu2017monotonic, fan2018online,arivazhagan2019}, GMM attention~\citep{graves2013generating, chiu2019comparison}, triggered attention (TA)~\citep{moritz2019triggered}, Scout Network~\citep{wang2020low}, and approaches that segment encoder output into non-overlapping chunks~\citep{jaitly2016neural, tsunoo2020streaming}. \citeauthor{tsunoo2020streaming} also applied knowledge distillation from the non-streaming model to the streaming model, but their streaming and non-streaming models do not share weights and are trained separately.

To improve the latency of RNN-T streaming models, \citeauthor{li2020towards} investigated additional early and late penalties on Endpointer prediction~\citep{chang2019joint} to reduce the emission latency, and employed the minimum word error rate (MWER) training~\citep{prabhavalkar2018minimum} to remedy accuracy degradation. \citeauthor{sainath2020streaming} further proposed to improve quality by using two-pass models~\citep{sainath2019two}, \ie, a second-pass LAS-based rescore model on top of the hypotheses from first-pass RNN-T streaming output. More recently, \citeauthor{li2020parallel} proposed parallel rescoring by replacing LSTMs with Transformers~\citep{vaswani2017attention} in rescoring models. \citeauthor{shuoyiin2020low} further proposed Prefetching to reduce system latency by submitting partial recognition results for subsequent processing such as obtaining assistant server responses or second-pass rescoring before the recognition result is finalized. Unlike these approaches, our work explores the unification of streaming and full-context ASR networks, thus can be generally applied as an add-on technique without requiring extra runtime support during inference.

\textbf{Weight Sharing for Multi-tasking.}
Sharing model weights of a deep neural network for multiple tasks has been widely explored in the literature to reduce overall model sizes. In the broadest sense, tasks can refer to different objectives or same objective but different settings, ranging from natural language processing and speech recognition to computer vision and reinforcement learning. In speech recognition, \citeauthor{kannan2019large} employed a single ASR network for multilingual ASR, and showed accuracy improvements over monolingual ASR systems. \citeauthor{wu2020dynamic} proposed dynamic sparsity neural networks (DSNN) for speech recognition on mobile devices with resource constraints. A single trained DSNN~\citep{wu2020dynamic} can transform into multiple networks of different sparsities for adaptive inference in real-time. \citeauthor{chang2019joint} trained a single RNN-T model with LSTMs~\citep{hochreiter1997long} for Joint Endpointing (\ie, predicting both recognition tokens and the end of an utterance transcription) in streaming ASR systems. Moreover, \citeauthor{watanabe2017hybrid} proposed a hybrid CTC and attention architecture for ASR based on multi-objective learning to eliminate the use of linguistic resources.

Another related research work in Computer Vision is Slimmable Neural Networks~\citep{yu2018slimmable, yu2019autoslim, yu2019universally, yu2020bignas}. \citeauthor{yu2018slimmable} proposed an approach to train a single neural network running at different widths, permitting instant and adaptive accuracy efficiency trade-offs at runtime. We also adapt the training rules introduced in slimmable networks, that is, using independent normalization layers for different sub-networks (tasks) as conditional parameters and using the prediction of teacher network to supervise student network as inplace distillation during the training. Unlike slimmable networks in which a large model is used to supervise a small model, we propose to distill the knowledge from full-context mode (teacher) into streaming mode (student) on the fly within the same Dual-mode ASR model.

\textbf{Knowledge Distillation.}
\citeauthor{hinton2015distilling} explored a simple method to ``transfer'' knowledge from a teacher neural network to a student neural network by enforcing their predictions to be close measured by KL-divergence, \(\ell_1\) or \(\ell_2\) distance. It is shown that such distillation method is effective to compress neural networks~\citep{yu2019universally}, accelerate training~\citep{chen2015net2net}, improve robustness~\citep{carlini2017towards, papernot2016distillation}, estimate model uncertainty~\citep{blundell2015weight} and transfer learned domain to other domains~\citep{tzeng2015simultaneous}.

%% file: sec3_approach.tex
\section{Dual-mode ASR}
Most neural sequence transduction networks for ASR have an encoder-decoder structure~\citep{graves2012sequence, sainath2020streaming, he2019streaming, li2020towards}, as shown in Figure~\ref{figs:overview_similarity}. Without loss of generality, here we discuss how to design Dual-mode ASR networks under the most commonly used RNN-T model~\citep{graves2012sequence}. In RNN-T models, we first extract mel-spectrogram feature from input speech waveform. The Mel-spectrogram feature is then fed into a neural-net encoder, which usually consists of feed-forward layers, RNN/LSTM layers, convolution layers, attention layers, pooling (time-reduction) layers, and residual or dense connections. In neural-net encoders, streaming ASR model requires all components to be auto-regressive, whereas full-context ASR model has no such requirement. The ASR decoder then predicts the token of current frame based on the output from the encoder and previous predicted tokens (inference) or target tokens (training with teacher forcing~\citep{williams1989learning}). The decoder is commonly an auto-regressive model in both streaming and full-context ASR models, thus is \textit{fully shared} in Dual-mode ASR. The prediction from decoder is finally used either in decoding algorithm during inference (\eg, beam search) or learning algorithm during training (\eg, RNN-T loss).

As discussed above and shown in Figure~\ref{figs:overview_similarity}, it becomes clear that the major difference between streaming and full-context ASR models is in the \textit{neural-net encoder}. In the following, we will first discuss the design principles of \textit{dual-mode encoder} to support both streaming and full-context ASR. We provide examples including dual-mode convolution, dual-mode average pooling, and dual-mode attention layers, which are widely used in the state-of-the-art ASR networks ContextNet~\citep{han2020contextnet} and Conformer~\citep{gulati2020conformer}. We will then discuss the training algorithm of Dual-mode ASR networks including joint training and inplace knowledge distillation.

\subsection{Dual-mode Encoder}
\label{secs:dual_mode_encoder}
Unifying streaming and full-context ASR models requires two design principles of \textit{Dual-mode Encoder}:
\begin{enumerate}[topsep=0pt, partopsep=0pt, itemsep=2pt, parsep=2pt]
    \item Each layer in a dual-mode encoder should be either dual-mode or streaming (\aka, causal). Since streaming encoder has to be auto-regressive which prohibits any future context, any full-context (\aka, non-causal) layer violates this constraint.
    \item The design of a dual-mode layer should not introduce significant amount of additional parameters, compared with its streaming model. We aim at supporting full-context ASR on top of the streaming model with near-zero parameters overhead.
\end{enumerate}
We show examples below by applying the above two design principles to ContextNet~\citep{han2020contextnet} and Conformer~\citep{gulati2020conformer}, in which the encoders are composed of pointwise operators (feed-forward net, residual connections, activation layers, striding, dropout, \etc), convolution, average pooling, self-attention and normalization layers.

\textbf{Pointwise operators are naturally dual-mode layers.} Neural network layers that connect input and output neurons within each timestep (no across-connections among different timesteps) are often referred as pointwise operators~\citep{chollet2017xception}, including feed-forward layers (\aka, fully-connected layers or \(1\times1\) convolution layers), activation layers (\eg, ReLU, Swish~\citep{ramachandran2017searching}), residual and dense connections~\citep{he2016deep, huang2017densely}, striding layers, dropout layers~\citep{srivastava2014dropout} and element-wise multiplications. As there is no information propagation through time, pointwise operators are naturally dual-mode layers and can be directly used in Dual-mode ASR encoders.

\begin{figure}[t]
\centering
\includegraphics[width=\linewidth]{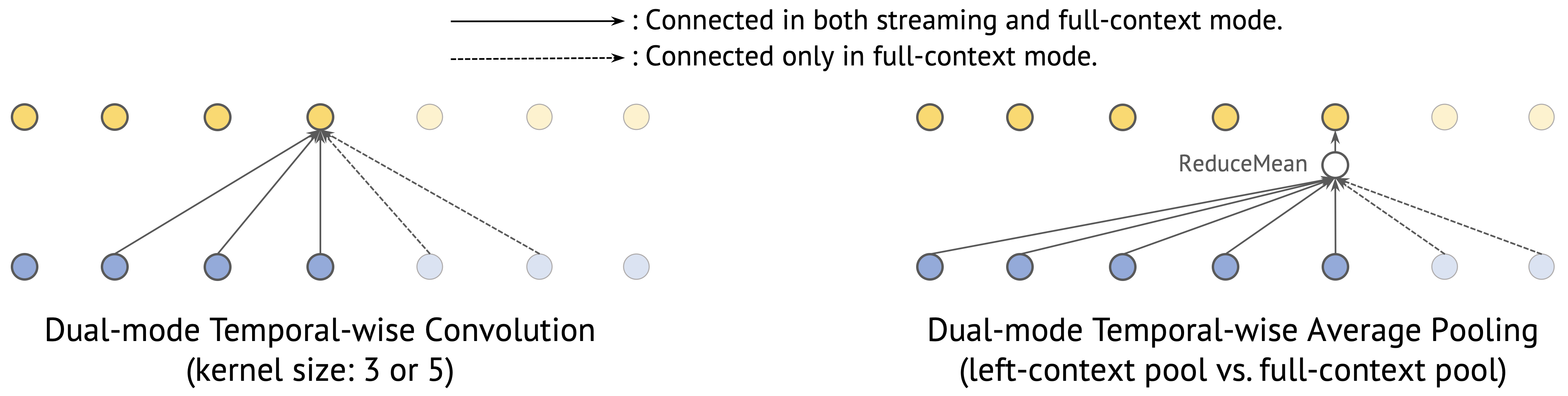}
\vspace{-0.5cm}
\caption{Dual-mode convolution and average pooling layer for Dual-mode ASR.}
\vspace{-0.5cm}
\label{figs:dual_mode_convolution}
\end{figure}

\textbf{Dual-mode Convolution.}
Convolution layers, however, convolve feature across its neighbor timesteps within a fixed window (\eg, kernel size is 3, 5, or larger), and has been widely used in sequence modeling~\citep{gehring2017convolutional, han2020contextnet, gulati2020conformer}. In conv-based streaming ASR models, causal convolution layers~\citep{oord2016wavenet} are used where the convolution window is biased to the left (self-included). As shown in Figure~\ref{figs:dual_mode_convolution} on the left, to support both streaming and full-context modes with shared weights, we first construct a normal symmetric convolution of kernel size \(k\) which will be applied in full-context mode. Then we mimic the causal convolution of kernel size \((k+1)/2\) by constructing a Boolean mask and multiplying with the full-context convolution kernel before applying the actual convolution of streaming mode in Dual-mode ASR encoders.

The design of dual-mode convolution introduces \((k-1)/2\) additional parameters to support full-context convolution (\(k\)) compared with streaming convolution (\((k+1)/2\)). However, we note that in convolution-based models, these temporal-wise convolution layers only take a tiny amount of total model size and most of the weights are on \(1\times1\) convolution layers which are fully shared pointwise operators. For example, in ContextNet~\citep{han2020contextnet}, temporal-wise convolution has less than 1\% of total model size, thus parameters overhead is negligible.

\textbf{Dual-mode Average Pooling.}
Squeeze-and-excitation~\citep{hu2018squeeze} (SE) modules are used in ContextNet to enhance the global context encoding. Each SE module is a sequential stack of average pooling (through time) layer, feed-forward layer, activation layer, another feed-forward layer and elementwise multiplication. To support both modes, dual-mode average pooling layer is used as shown in Figure~\ref{figs:dual_mode_convolution} on the right. Dual-mode average pooling layer is parameter-free thus does not introduce additional model parameters. It also trains in parallel in streaming mode, easily implemented with ``cumsum'' function in both TensorFlow and PyTorch.

\begin{wrapfigure}{r}{0.5\textwidth}
\centering
\includegraphics[width=0.98\linewidth]{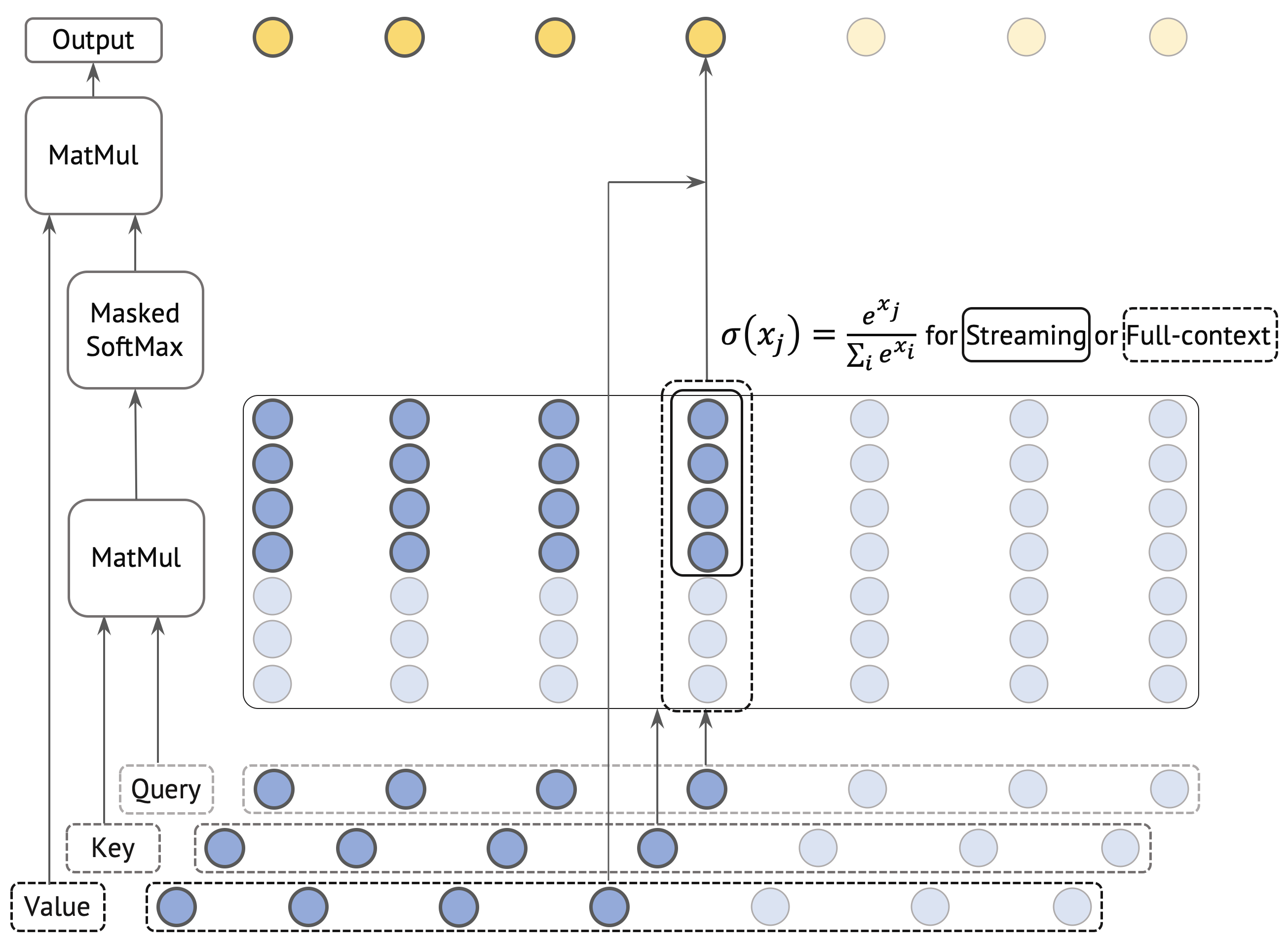}
\vspace{-0.3cm}
\caption{Dual-mode self-attention layer.}
\vspace{-0.3cm}
\label{figs:dual_mode_attention}
\end{wrapfigure}
\textbf{Dual-mode Self-attention.}
Self-attention (\aka~intra-attention) is an attention mechanism weighting different positions of a single sequence in order to compute a representation of the same sequence. It is heavily used in Conformer~\citep{gulati2020conformer} ASR networks. The attention layer itself is parameter-free (projection layers before attention are fully shared), and is composed of matrix multiplication of the key and the query, followed by softmax over keys, before another matrix multiplication with the value. As shown in Figure~\ref{figs:dual_mode_attention}, in dual-mode attention layer, the softmax is performed on the left context only in streaming mode (rectangle with solid line), compared with the full-context mode (rectangle with dash line). We find this simple form of dual-mode self-attention works well in practice.

\textbf{Dual-mode Normalization.} Moreover, following~\cite{yu2018slimmable}, we also find the normalization statistics like means and variances are different in streaming and full-context modes. Thus, for normalization layers including BatchNorm~\citep{ioffe2015batch} and LayerNorm~\citep{ba2016layer} in Dual-mode ContextNet and Dual-mode Conformer, we instantiate two separate norm layers dedicated to streaming and full-context mode respectively.

\subsection{Training Dual-mode ASR Networks}
The training algorithm of Dual-mode ASR networks is outlined in Algorithm~\ref{algos:algo}. In this section, we discuss two important training techniques: joint training and inplace knowledge distillation.

\textbf{Joint Training.}
To train Dual-mode ASR networks, given a batch of data in each training iteration, we can either randomly sample one from two modes to train, or train both modes and aggregate their losses. In the former approach, referred as \textit{randomly sampled training}, we can control the importance of streaming and full-context modes by setting different sampling probabilities during training. In the latter approach, referred as \textit{joint training}, importance can also be controlled by assigning different loss weights to balance streaming and full-context modes. Empirically we find joint training leads to better model qualities overall thus is adopted in all of our experiments. We will show an ablation study comparing randomly sampled training and joint training. In all of our experiments, we treat streaming and full-context mode to be equally important by assigning equal importance during training.

\textbf{Inplace Knowledge Distillation.}
Additionally we propose to distill knowledge from the full-context mode (teacher) into the streaming mode (student) on the fly within the same Dual-mode ASR model, by encouraging consistency of the predicted token probabilities. Since in each iteration we always compute predictions of both modes, the teacher prediction comes for free (no additional computation or memory cost), as shown in Algorithm~\ref{algos:algo}. We use the efficient knowledge distillation introduced by \citeauthor{panchapagesan2020efficient}, which is based on the KL-divergence between full-context and streaming over the probability of three parts: \(P_{label}\), \(P_{blank}\) and \(1 - P_{label} - P_{blank}\). We note that the prediction of full-context mode (teacher) usually has lower latency (since it has no incentive to delay its output), thus we can control the target emission latency of streaming mode (student) by shifting the prediction of full-context mode, before applying distillation loss. We do a small-scale hyper-parameter sweep from -2 to 2 frames to shift for ContextNet and Conformer in our experiments.

\input{algos/algorithm}

%% file: algos/algorithm.tex
\begin{algorithm}[t]
\caption{Pseudocode of training Dual-mode ASR networks.}
\label{algos:algo}
\definecolor{codeblue}{rgb}{0.25,0.5,0.5}
\lstset{
  backgroundcolor=\color{white},
  basicstyle=\fontsize{7.2pt}{7.2pt}\ttfamily\selectfont,
  columns=fullflexible,
  breaklines=true,
  captionpos=b,
  commentstyle=\fontsize{7.2pt}{7.2pt}\color{codeblue},
  keywordstyle=\fontsize{7.2pt}{7.2pt},
}
\begin{lstlisting}[language=python]
# Requires: data_loader; context manager with support of mode switching by network.mode(); dual_mode_network with support of running both modes under context manager;

for x, y in data_loader:  # Load a minibatch of speech input x and text label y.
    with dual_mode_network.mode('fullcontext'):  # Switch context to 'fullcontext' mode.
        # Compute full-context prediction given speech input x and text label y.
        fullcontext_pred = dual_mode_network.forward_encoder_decoder(x, y)
        # Compute RNN-T loss of full-context mode.
        fullcontext_loss = rnnt_loss(fullcontext_pred, y)
        
    with dual_mode_network.mode('streaming'):  # Switch context to 'streaming' mode.
        # Compute streaming prediction given speech input x and text label y.
        streaming_pred = dual_mode_network.forward_encoder_decoder(x, y)
        # Compute RNN-T loss of streaming mode.
        streaming_loss = rnnt_loss(streaming_pred, y)
    
    # Add inplace knowledge distillation loss (full-context prediction as teacher).
    distill_loss = inplace_distill_loss(streaming_pred, stop_gradient(fullcontext_pred))
        
    # Compute total loss as a sum of full-context, streaming and distillation losses.
    loss = fullcontext_loss + streaming_loss + distill_loss
    loss.backward()  # Update weights.
\end{lstlisting}
\end{algorithm}

%% file: sec4_experiments.tex
\section{Experiments}

\subsection{Main Result}
\textbf{Measuring Latency.}
Latency measurement is itself challenging for streaming ASR systems. Motivated by Prefetching~\citep{shuoyiin2020low} technique, we measure latency as the difference of two timestamps: 1) when the last token is emitted in the finalized recognition result; 2) the end of the speech when a user finishes speaking. We find this is especially descriptive of user experience in real-world ASR applications like Voice Search. ASR models that capture stronger contexts can emit the full hypothesis even before they are spoken, leading to a \textbf{negative latency}. Moreover, instead of naively averaging latency over all utterances, we report both median and 90th percentile of all utterances in test set, denoted as \textbf{Latency@50} and \textbf{Latency@90}, to better characterize latency by excluding outlier utterances. To evaluate the model quality, we report WER only for full-context models and both WER and latency for streaming models (full-context latency is meaningless).

\textbf{Datasets.}
We conduct our experiments on two datasets: a public widely used dataset LibriSpeech~\citep{panayotov2015librispeech} (1,000 hours of English reading speech) and a large-scale dataset MultiDomain (413,000 hours speech, 287 million utterances of a mixture across multiple domains including Voice Search, YouTube, and Meetings). Table~\ref{tabs:dataset} summarizes the information and statistics of two datasets. For LibriSpeech, we report our evaluation results on TestClean and TestOther (noisy) sets and compare with other published baselines. For MultiDomain, we report our evaluation results on Voice Search test set and compare with our reproduced baselines. For fair comparisons, on each dataset we train and report our models and baselines with the same settings (number of training iterations, hyper-parameters, optimizer, regularization, \etc). We note that these hyper-parameters are inherited from previous work~\cite{han2020contextnet, gulati2020conformer} and not specifically tuned for our dual-mode models.

\input{tabs/dataset}

\textbf{ASR Networks.}
We use two recent state-of-the-art ASR networks to demonstrate the effectiveness of our proposed methods, ContextNet~\citep{han2020contextnet} and Conformer~\citep{gulati2020conformer}. The encoder of ContextNet is based on depthwise-separable convolution~\citep{chollet2017xception} and squeeze-and-excitation modules~\citep{hu2018squeeze}. In depthwise-separable convolution of Dual-mode ContextNet, the weights of \(1\times1\) convolutions are fully shared between streaming and full-context mode, whereas for temporal-wise convolution we follow the design of Dual-mode Convolution proposed in Section~\ref{secs:dual_mode_encoder}. Note that in ContextNet, temporal-wise convolutions only take less than 1\% of the model size thus the parameters overhead of full-context mode is negligible compared with steaming mode. In squeeze-and-excitation modules, we use dual-mode average pooling layers (Section~\ref{secs:dual_mode_encoder}) to support both streaming and full-context mode without additional parameters. 

Conformer~\citep{gulati2020conformer} combines convolution and transformer to model both local and global dependencies of speech sequences in a parameter-efficient way. In Dual-mode Conformer, we replace all convolution and transformer layers with their dual-mode correspondents (Section~\ref{secs:dual_mode_encoder}). Moreover, for normalization layers including BatchNorm~\citep{ioffe2015batch} and LayerNorm~\citep{ba2016layer} in Dual-mode ContextNet and Dual-mode Conformer, we instantiate two separate norm layers for streaming and full-context mode respectively.

\input{tabs/main_result}

\textbf{Training Details and Results.} We train our models exactly following our baselines ContextNet~\citep{han2020contextnet} and Conformer~\citep{gulati2020conformer}, using Adam optimizer~\citep{kingma2014adam}, SpecAugment~\citep{park2019specaugment} and a transformer learning rate schedule~\citep{vaswani2017attention} with warm-up~\citep{goyal2017accurate}. Our main results are summarized in Table~\ref{tabs:main_result_multidomain} and Table~\ref{tabs:main_result_librispeech}. We also add a streaming ContextNet Look-ahead baseline (6 frames, 10ms per frame, totally 60ms look-ahead latency) in Table~\ref{tabs:main_result_librispeech}  by padding additional frames at the end of the input utterances. As shown in the tables, the streaming mode in Dual-mode ASR models has significantly better latency and similar or higher WER results, surpassing other baselines including conventional models, LSTM-based transducers~\citep{sainath2020streaming}, transformer-transducers~\citep{zhang2020transformer} and some others.

\subsection{Ablation Study}
In this section, we perform various ablation studies to support and understand the effectiveness of each technique in Dual-mode ASR. We train Dual-mode ContextNet on LibriSpeech training set with exactly same settings and report WER, Latency@50 and Latency@90 on TestOther set of streaming mode. We specifically study three techniques and their combinations including \textit{weight sharing}, \textit{joint training} and \textit{inplace knowledge distillation} during the training. 

During the training we distill knowledge from full-context mode (teacher) into streaming mode (student) on the fly within the same dual-mode model. Inplace distillation during the training comes for free as shown in training Algorithm~\ref{algos:algo}. But what if we simply share weights and jointly train them without distillation? As shown in the second row of Table~\ref{tabs:ablation}, the model without inplace distillation during the training has worse results compared to the baseline.

Given a batch of data for each training iteration, we train both modes and aggregate their losses. We also show results of \textit{randomly sampled training} in the third row of Table~\ref{tabs:ablation}, which leads to even worse performance. Note that with randomly sampled training, we cannot apply inplace distillation easily either because in each training iteration there is only one prediction from either streaming mode or full-context mode.

Weight sharing reduces the model size which is one of the major motivation of Dual-mode ASR. However, what if we simply train two individual models and use knowledge distillation with full-context model as the teacher? As shown in the last row of Table~\ref{tabs:ablation}, the results are better than other ablation but still worse than the Dual-mode ASR baseline. It might indicate that weight sharing itself encourages learning better deep representation for streaming ASR. Weight sharing has been shown empirically to improve Multilingual ASR~\citep{kannan2019large}, Model Pruning~\citep{wu2020dynamic}, Endpointing~\citep{hochreiter1997long} and some Computer Vision problems~\citep{yu2018slimmable} and this intriguing property need to be studied in more details as a future work.

\input{tabs/ablation}

\begin{figure}[h]
\centering
\includegraphics[width=\linewidth]{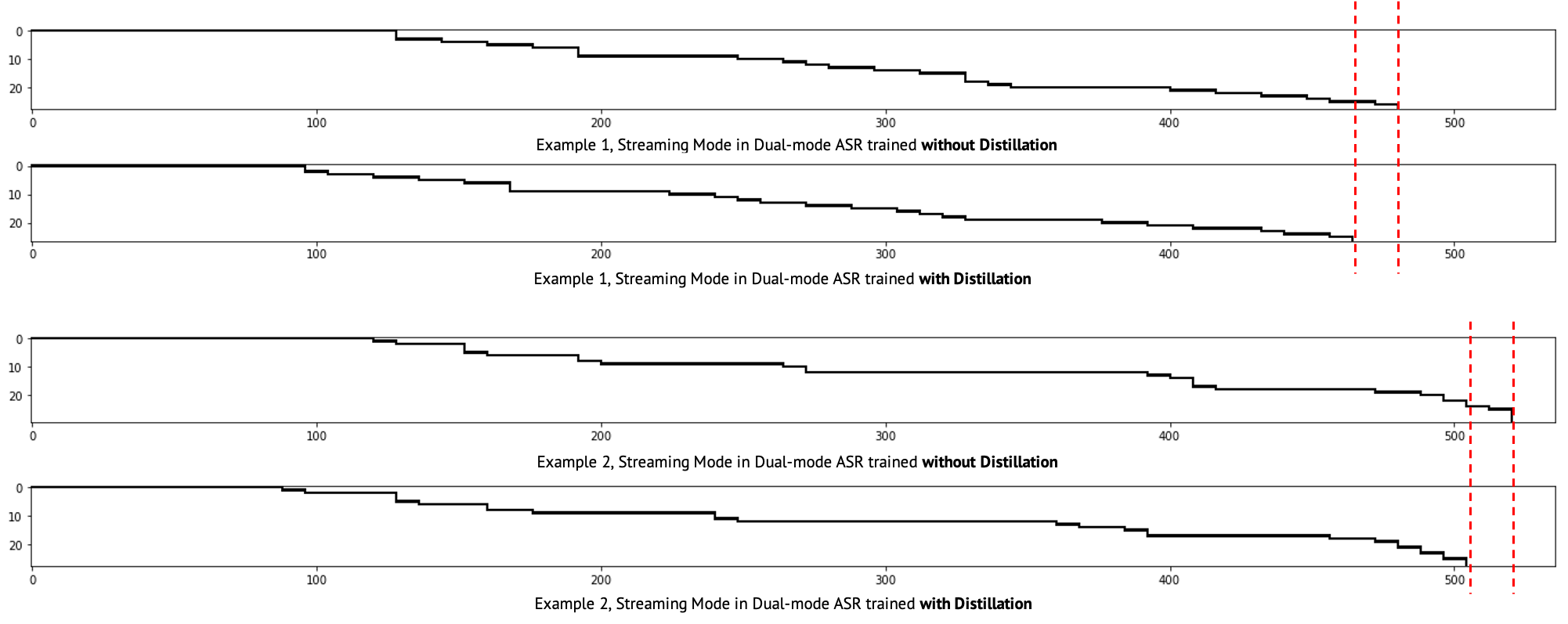}
\caption{Two speech-text pair comparison of Dual-model ASR models trained with and without inplace distillation by visualization of their streaming emission lattices. X-axis represents the speech input frames while Y-axis represents the text output labels (tokens). Inplace distillation significantly reduces emission latency of streaming mode in Dual-mode ASR models which is critical in real-world applications.}
\label{figs:distill_vis}
\end{figure}

Further, we visualize the emission lattices of dual-mode ASR models trained with and without inplace knowledge distillation. We randomly sampled two audio sequences on LibriSpeech TestOther set and plotted their emission lattices of streaming mode in Figure~\ref{figs:distill_vis}. X-axis represents the speech input frames while Y-axis represents the text output labels (tokens). Figure~\ref{figs:distill_vis} shows that with knowledge distillation from full-context mode in Dual-mode ASR, streaming mode emits faster and has much less latency, which is very critical for product datasets like MultiDomain presented in our work.

%% file: tabs/dataset.tex
\begin{table*}[ht]
\vspace{-0.3cm}
\caption{Summary of datasets we used in our experiments.}
\label{tabs:dataset}
\small
\centering
\begin{tabular}{@{}l l l l@{}}
\toprule
\textbf{Dataset Name} & \textbf{\# Hours} & \textbf{\# Utterances} & \textbf{Speech Domain}  \\
\midrule
\multirow{2}{*}{\shortstack[l]{LibriSpeech\\\citep{panayotov2015librispeech}}} & \multirow{2}{*}{\(\sim970\)} & \multirow{2}{*}{\(\sim281,000\)} & \multirow{2}{*}{Single domain of English reading speech.} \vspace{0.5cm}\\
\multirow{2}{*}{\shortstack[l]{MultiDomain\\\citep{narayanan2018toward}}} & \multirow{2}{*}{\(\sim413,000\)} & \multirow{2}{*}{\(\sim287,000,000\)} & \multirow{2}{*}{\shortstack[l]{Multiple domains including: Voice Search,\\ Farfield Speech,  YouTube and Meetings.}}\\
 & & & \\
\bottomrule
\end{tabular}
\end{table*}

%% file: tabs/main_result.tex
\begin{table*}[ht]
\vspace{-0.3cm}
\caption{Summary of our results on MultiDomain dataset~\citep{narayanan2018toward}. We report WER on Voice Search test set. Compared with standalone ContextNet and Conformer models, Dual-mode ASR models have slightly higher accuracy and much better streaming latency. ASR models that capture stronger contexts can emit the full hypothesis even slightly before they are spoken, leading to a \textit{negative latency}.}
\label{tabs:main_result_multidomain}
\centering
\small
\begin{tabular}{@{}l l c l l l@{}} \toprule
\textbf{Method} & \textbf{Mode} & \textbf{\# Params (M)} & \textbf{VS Test} & \textbf{Latency@50} & \textbf{Latency@90}\\
 &  &  & WER(\%) & (ms) & (ms)\\
\midrule
ContextNet & Full-context & 133 & 5.1 & ------ & ------ \\
Conformer & Full-context & 142 & 5.2 & ------ & ------ \\

\midrule
LSTM~\citep{sainath2020streaming} & Streaming & 179 & 6.4 & 190 & 350 \\
ContextNet~\citep{han2020contextnet} & Streaming & 133 & 6.1 & 160 & 310 \\
Conformer~\citep{gulati2020conformer} & Streaming & 142 & 6.1 & 160 & 300 \\
\midrule
\multirow{2}{*}{Dual-mode ContextNet} & Full-context & \multirow{2}{*}{133}  & 4.9 & ------ & ------ \\
& Streaming &  & 6.0 \textbf{\textsubscript{\textcolor{darkergreen}{(-0.1)}}} & 10 \textbf{\textsubscript{\textcolor{darkergreen}{(-150)}}} & 220 \textbf{\textsubscript{\textcolor{darkergreen}{(-90)}}} \vspace{0.2cm}\\
\multirow{2}{*}{Dual-mode Conformer} & Full-context & \multirow{2}{*}{142} & 5.0 & ------ & ------ \\
& Streaming &  & 6.0 \textbf{\textsubscript{\textcolor{darkergreen}{(-0.1)}}} & -50 \textbf{\textsubscript{\textcolor{darkergreen}{(-210)}}} & 130 \textbf{\textsubscript{\textcolor{darkergreen}{(-170)}}}\\
\bottomrule
\end{tabular}
\end{table*}

\begin{table*}[ht]
\vspace{-0.3cm}
\caption{Summary of our results on Librispeech dataset~\citep{panayotov2015librispeech}. We report WER on TestClean and TestOther (noisy) set. Compared with standalone ContextNet and Conformer models, Dual-mode ASR models have both higher accuracy in average and better streaming latency.}
\label{tabs:main_result_librispeech}
\centering
\small
\begin{tabular}{@{}l l c l l l@{}} \toprule
\textbf{Method} & \textbf{Mode} & \textbf{\# Params (M)} &\textbf{Test Clean/Other} & \textbf{Latency@50} & \textbf{Latency@90}\\
& & & WER(\%) & (ms) & (ms)\\
\midrule
LSTM-LAS & Full-context & 360 & 2.6 \hspace{0.2cm}/\hspace{0.2cm} 6.0 & ------ & ------ \\
QuartzNet-CTC & Full-context & 19 & 3.9 \hspace{0.2cm}/\hspace{0.2cm} 11.3 & ------ & ------ \\
Transformer & Full-context & 29 & 3.1 \hspace{0.2cm}/\hspace{0.2cm} 7.3 & ------ & ------ \\
Transformer & Full-context & 139 & 2.4 \hspace{0.2cm}/\hspace{0.2cm} 5.6 & ------ & ------ \\
ContextNet & Full-context & 31.4 & 2.4 \hspace{0.2cm}/\hspace{0.2cm} 5.4 & ------ & ------ \\
Conformer & Full-context & 30.7 & 2.3 \hspace{0.2cm}/\hspace{0.2cm} 5.0 & ------ & ------ \\

\midrule
Transformer & Streaming & 18.9 & 5.0 \hspace{0.2cm}/\hspace{0.1cm} 11.6 & 80 & 190 \\
ContextNet & Streaming & 31.4 & 4.5 \hspace{0.2cm}/\hspace{0.1cm} 10.0 & 70 & 270 \\
Conformer & Streaming & 30.7 & 4.6 \hspace{0.2cm}/\hspace{0.2cm} 9.9 & 140 & 280 \\
ContextNet Look-ahead & Streaming & 31.4 & 4.1 \hspace{0.2cm}/\hspace{0.2cm} 9.0 & 150 & 420 \\

\midrule
\multirow{2}{*}{Dual-mode Transformer} & Full-context & \multirow{2}{*}{29} & 3.1 \hspace{0.2cm}/\hspace{0.2cm} 7.9 & ------ & ------ \\
& Streaming &  &  4.4 \textbf{\textsubscript{\textcolor{darkergreen}{(-0.6)}}} \hspace{0cm}/\hspace{0cm} 11.5 \textbf{\textsubscript{\textcolor{darkergreen}{(-0.1)}}} & -50  \textbf{\textsubscript{\textcolor{darkergreen}{(-130)}}} &  30 \textbf{\textsubscript{\textcolor{darkergreen}{(-160)}}} \vspace{0.2cm}\\

\multirow{2}{*}{Dual-mode ContextNet} & Full-context & \multirow{2}{*}{31.8} & 2.3 \hspace{0.2cm}/\hspace{0.2cm} 5.3 & ------ & ------ \\
& Streaming &  & 3.9 \textbf{\textsubscript{\textcolor{darkergreen}{(-0.6)}}} \hspace{0cm}/\hspace{0cm} 8.5 \textbf{\textsubscript{\textcolor{darkergreen}{(-1.5)}}} & 40 \textbf{\textsubscript{\textcolor{darkergreen}{(-30)}}} & 160 \textbf{\textsubscript{\textcolor{darkergreen}{(-110)}}} \vspace{0.2cm}\\
\multirow{2}{*}{Dual-mode Conformer} & Full-context & \multirow{2}{*}{30.7} & 2.5 \hspace{0.2cm}/\hspace{0.2cm} 5.9 & ------ & ------ \\
& Streaming &  & 3.7 \textbf{\textsubscript{\textcolor{darkergreen}{(-0.9)}}} \hspace{0cm}/\hspace{0cm} 9.2 \textbf{\textsubscript{\textcolor{darkergreen}{(-0.7)}}} & 10 \textbf{\textsubscript{\textcolor{darkergreen}{(-130)}}} & 90 \textbf{\textsubscript{\textcolor{darkergreen}{(-190)}}}  \\
\bottomrule
\end{tabular}
\end{table*}

%% file: tabs/ablation.tex
\begin{table*}[ht]
\vspace{-0.3cm}
\caption{Ablation studies of weight sharing, joint training and inplace distillation. We report WER on TestOther (noisy) set~\citep{panayotov2015librispeech} using ContextNet with same training settings.}
\label{tabs:ablation}
\small
\centering
\begin{tabular}{@{}c c c l l l@{}}
\toprule
\textbf{Weight Sharing} & \textbf{Joint Training} & \textbf{Inplace Distillation} & \textbf{TestOther} & \textbf{Latency@50} & \textbf{Latency@90} \\
 &  &  & WER(\%) & (ms) & (ms) \\
\midrule
 \textcolor{darkergreen}{\ding{52}} & \textcolor{darkergreen}{\ding{52}} & \textcolor{darkergreen}{\ding{52}} & 8.5 & 40 & 160 \\
 \textcolor{darkergreen}{\ding{52}} & \textcolor{darkergreen}{\ding{52}} & \textcolor{red2}{\ding{56}} & 10.2 \textbf{\textsubscript{\textcolor{red2}{(+1.7)}}} & 120  \textbf{\textsubscript{\textcolor{red2}{(+80)}}}& 310  \textbf{\textsubscript{\textcolor{red2}{(+150)}}} \\ 
 \textcolor{darkergreen}{\ding{52}} & \textcolor{red2}{\ding{56}} & \textcolor{red2}{\ding{56}} & 10.6 \textbf{\textsubscript{\textcolor{red2}{(+2.1)}}} & 90 \textbf{\textsubscript{\textcolor{red2}{(+50)}}}& 290 \textbf{\textsubscript{\textcolor{red2}{(+130)}}} \\
 \textcolor{red2}{\ding{56}} & \textcolor{darkergreen}{\ding{52}} & \textcolor{darkergreen}{\ding{52}} & 9.9 \textbf{\textsubscript{\textcolor{red2}{(+1.4)}}}& 50 \textbf{\textsubscript{\textcolor{red2}{(+10)}}}& 210 \textbf{\textsubscript{\textcolor{red2}{(+50)}}}\\
\bottomrule
\end{tabular}
\end{table*}

%% file: sec5_conclusion.tex
\section{Conclusion}
In this work, we have proposed a unified framework, \textit{Dual-mode ASR}, to unify and improve streaming ASR by joint full-context modeling.
We hope our exploration will inspire streaming models in other fields such as simultaneous machine translation and video processing.

%% file: main.bbl
\begin{thebibliography}{68}
\providecommand{\natexlab}[1]{#1}
\providecommand{\url}[1]{\texttt{#1}}
\expandafter\ifx\csname urlstyle\endcsname\relax
  \providecommand{\doi}[1]{doi: #1}\else
  \providecommand{\doi}{doi: \begingroup \urlstyle{rm}\Url}\fi

\bibitem[Arivazhagan et~al.(2019)Arivazhagan, Cherry, Macherey, Chiu, Yavuz,
  Pang, Li, and Raffel]{arivazhagan2019}
Naveen Arivazhagan, Colin Cherry, Wolfgang Macherey, Chung-Cheng Chiu, Semih
  Yavuz, Ruoming Pang, Wei Li, and Colin Raffel.
\newblock Monotonic infinite lookback attention for simultaneous machine
  translation.
\newblock In \emph{ACL}, 2019.

\bibitem[Ba et~al.(2016)Ba, Kiros, and Hinton]{ba2016layer}
Jimmy~Lei Ba, Jamie~Ryan Kiros, and Geoffrey~E Hinton.
\newblock Layer normalization.
\newblock \emph{arXiv preprint arXiv:1607.06450}, 2016.

\bibitem[Bahdanau et~al.(2016)Bahdanau, Chorowski, Serdyuk, Brakel, and
  Bengio]{bahdanau2016end}
Dzmitry Bahdanau, Jan Chorowski, Dmitriy Serdyuk, Philemon Brakel, and Yoshua
  Bengio.
\newblock End-to-end attention-based large vocabulary speech recognition.
\newblock In \emph{2016 IEEE international conference on acoustics, speech and
  signal processing (ICASSP)}, pp.\  4945--4949. IEEE, 2016.

\bibitem[Blundell et~al.(2015)Blundell, Cornebise, Kavukcuoglu, and
  Wierstra]{blundell2015weight}
Charles Blundell, Julien Cornebise, Koray Kavukcuoglu, and Daan Wierstra.
\newblock Weight uncertainty in neural networks.
\newblock \emph{arXiv preprint arXiv:1505.05424}, 2015.

\bibitem[Carlini \& Wagner(2017)Carlini and Wagner]{carlini2017towards}
Nicholas Carlini and David Wagner.
\newblock Towards evaluating the robustness of neural networks.
\newblock In \emph{2017 ieee symposium on security and privacy (sp)}, pp.\
  39--57. IEEE, 2017.

\bibitem[Chan et~al.(2016)Chan, Jaitly, Le, and Vinyals]{chan2016listen}
William Chan, Navdeep Jaitly, Quoc Le, and Oriol Vinyals.
\newblock Listen, attend and spell: A neural network for large vocabulary
  conversational speech recognition.
\newblock In \emph{2016 IEEE International Conference on Acoustics, Speech and
  Signal Processing (ICASSP)}, pp.\  4960--4964. IEEE, 2016.

\bibitem[Chang et~al.(2019)Chang, Prabhavalkar, He, Sainath, and
  Simko]{chang2019joint}
Shuo-Yiin Chang, Rohit Prabhavalkar, Yanzhang He, Tara~N Sainath, and Gabor
  Simko.
\newblock Joint endpointing and decoding with end-to-end models.
\newblock In \emph{ICASSP 2019-2019 IEEE International Conference on Acoustics,
  Speech and Signal Processing (ICASSP)}, pp.\  5626--5630. IEEE, 2019.

\bibitem[Chang et~al.(2020)Chang, Li, Rybach, He, Li, Sainath, and
  Strohman]{shuoyiin2020low}
Shuo-Yiin Chang, Bo~Li, David Rybach, Yanzhang He, Wei Li, Tara Sainath, and
  Trevor Strohman.
\newblock Low latency speech recognition using end-to-end prefetching.
\newblock In \emph{Interspeech}. ISCA, 2020.

\bibitem[Chen et~al.(2015)Chen, Goodfellow, and Shlens]{chen2015net2net}
Tianqi Chen, Ian Goodfellow, and Jonathon Shlens.
\newblock Net2net: Accelerating learning via knowledge transfer.
\newblock \emph{arXiv preprint arXiv:1511.05641}, 2015.

\bibitem[Chiu \& Raffel(2017)Chiu and Raffel]{chiu2017monotonic}
Chung-Cheng Chiu and Colin Raffel.
\newblock Monotonic chunkwise attention.
\newblock \emph{arXiv preprint arXiv:1712.05382}, 2017.

\bibitem[Chiu \& Raffel(2018)Chiu and Raffel]{chiu2018monotonic}
Chung-Cheng Chiu and Colin Raffel.
\newblock Monotonic chunkwise attention.
\newblock In \emph{International Conference on Learning Representations}, 2018.

\bibitem[Chiu et~al.(2019)Chiu, Han, Zhang, Pang, Kishchenko, Nguyen,
  Narayanan, Liao, Zhang, Kannan, Prabhavalkar, Chen, Sainath, and
  Wu]{chiu2019comparison}
Chung-Cheng Chiu, Wei Han, Yu~Zhang, Ruoming Pang, Sergey Kishchenko, Patrick
  Nguyen, Arun Narayanan, Hank Liao, Shuyuan Zhang, Anjuli Kannan, Rohit
  Prabhavalkar, Zhifeng Chen, Tara Sainath, and Yonghui Wu.
\newblock A comparison of end-to-end models for long-form speech recognition.
\newblock In \emph{ASRU}, 2019.

\bibitem[Chollet(2017)]{chollet2017xception}
Fran{\c{c}}ois Chollet.
\newblock Xception: Deep learning with depthwise separable convolutions.
\newblock In \emph{Proceedings of the IEEE conference on computer vision and
  pattern recognition}, pp.\  1251--1258, 2017.

\bibitem[Chorowski \& Jaitly(2016)Chorowski and Jaitly]{chorowski2016towards}
Jan Chorowski and Navdeep Jaitly.
\newblock Towards better decoding and language model integration in sequence to
  sequence models.
\newblock \emph{arXiv preprint arXiv:1612.02695}, 2016.

\bibitem[Chorowski et~al.(2014)Chorowski, Bahdanau, Cho, and
  Bengio]{chorowski2014end}
Jan Chorowski, Dzmitry Bahdanau, Kyunghyun Cho, and Yoshua Bengio.
\newblock End-to-end continuous speech recognition using attention-based
  recurrent nn: First results.
\newblock \emph{arXiv preprint arXiv:1412.1602}, 2014.

\bibitem[Chorowski et~al.(2015)Chorowski, Bahdanau, Serdyuk, Cho, and
  Bengio]{chorowski2015attention}
Jan~K Chorowski, Dzmitry Bahdanau, Dmitriy Serdyuk, Kyunghyun Cho, and Yoshua
  Bengio.
\newblock Attention-based models for speech recognition.
\newblock In \emph{Advances in neural information processing systems}, pp.\
  577--585, 2015.

\bibitem[Fan et~al.(2018)Fan, Zhou, Chen, Jia, and Liu]{fan2018online}
Ruchao Fan, Pan Zhou, Wei Chen, Jia Jia, and Gang Liu.
\newblock An online attention-based model for speech recognition.
\newblock \emph{arXiv preprint arXiv:1811.05247}, 2018.

\bibitem[Gehring et~al.(2017)Gehring, Auli, Grangier, Yarats, and
  Dauphin]{gehring2017convolutional}
Jonas Gehring, Michael Auli, David Grangier, Denis Yarats, and Yann~N Dauphin.
\newblock Convolutional sequence to sequence learning.
\newblock \emph{arXiv preprint arXiv:1705.03122}, 2017.

\bibitem[Goyal et~al.(2017)Goyal, Doll{\'a}r, Girshick, Noordhuis, Wesolowski,
  Kyrola, Tulloch, Jia, and He]{goyal2017accurate}
Priya Goyal, Piotr Doll{\'a}r, Ross Girshick, Pieter Noordhuis, Lukasz
  Wesolowski, Aapo Kyrola, Andrew Tulloch, Yangqing Jia, and Kaiming He.
\newblock Accurate, large minibatch sgd: Training imagenet in 1 hour.
\newblock \emph{arXiv preprint arXiv:1706.02677}, 2017.

\bibitem[Graves(2012)]{graves2012sequence}
Alex Graves.
\newblock Sequence transduction with recurrent neural networks.
\newblock \emph{arXiv preprint arXiv:1211.3711}, 2012.

\bibitem[Graves(2013)]{graves2013generating}
Alex Graves.
\newblock Generating sequences with recurrent neural networks, 2013.

\bibitem[Gulati et~al.(2020)Gulati, Qin, Chiu, Parmar, Zhang, Yu, Han, Wang,
  Zhang, Wu, et~al.]{gulati2020conformer}
Anmol Gulati, James Qin, Chung-Cheng Chiu, Niki Parmar, Yu~Zhang, Jiahui Yu,
  Wei Han, Shibo Wang, Zhengdong Zhang, Yonghui Wu, et~al.
\newblock Conformer: Convolution-augmented transformer for speech recognition.
\newblock \emph{arXiv preprint arXiv:2005.08100}, 2020.

\bibitem[Han et~al.(2020)Han, Zhang, Zhang, Yu, Chiu, Qin, Gulati, Pang, and
  Wu]{han2020contextnet}
Wei Han, Zhengdong Zhang, Yu~Zhang, Jiahui Yu, Chung-Cheng Chiu, James Qin,
  Anmol Gulati, Ruoming Pang, and Yonghui Wu.
\newblock Contextnet: Improving convolutional neural networks for automatic
  speech recognition with global context.
\newblock \emph{arXiv preprint arXiv:2005.03191}, 2020.

\bibitem[He et~al.(2016)He, Zhang, Ren, and Sun]{he2016deep}
Kaiming He, Xiangyu Zhang, Shaoqing Ren, and Jian Sun.
\newblock Deep residual learning for image recognition.
\newblock In \emph{Proceedings of the IEEE conference on computer vision and
  pattern recognition}, pp.\  770--778, 2016.

\bibitem[He et~al.(2019)He, Sainath, Prabhavalkar, McGraw, Alvarez, Zhao,
  Rybach, Kannan, Wu, Pang, et~al.]{he2019streaming}
Yanzhang He, Tara~N Sainath, Rohit Prabhavalkar, Ian McGraw, Raziel Alvarez,
  Ding Zhao, David Rybach, Anjuli Kannan, Yonghui Wu, Ruoming Pang, et~al.
\newblock Streaming end-to-end speech recognition for mobile devices.
\newblock In \emph{ICASSP 2019-2019 IEEE International Conference on Acoustics,
  Speech and Signal Processing (ICASSP)}, pp.\  6381--6385. IEEE, 2019.

\bibitem[Hinton et~al.(2015)Hinton, Vinyals, and Dean]{hinton2015distilling}
Geoffrey Hinton, Oriol Vinyals, and Jeff Dean.
\newblock Distilling the knowledge in a neural network.
\newblock \emph{arXiv preprint arXiv:1503.02531}, 2015.

\bibitem[Hochreiter \& Schmidhuber(1997)Hochreiter and
  Schmidhuber]{hochreiter1997long}
Sepp Hochreiter and J{\"u}rgen Schmidhuber.
\newblock Long short-term memory.
\newblock \emph{Neural computation}, 9\penalty0 (8):\penalty0 1735--1780, 1997.

\bibitem[Hu et~al.(2018)Hu, Shen, and Sun]{hu2018squeeze}
Jie Hu, Li~Shen, and Gang Sun.
\newblock Squeeze-and-excitation networks.
\newblock In \emph{Proceedings of the IEEE conference on computer vision and
  pattern recognition}, pp.\  7132--7141, 2018.

\bibitem[Huang et~al.(2017)Huang, Liu, Van Der~Maaten, and
  Weinberger]{huang2017densely}
Gao Huang, Zhuang Liu, Laurens Van Der~Maaten, and Kilian~Q Weinberger.
\newblock Densely connected convolutional networks.
\newblock In \emph{Proceedings of the IEEE conference on computer vision and
  pattern recognition}, pp.\  4700--4708, 2017.

\bibitem[Huang et~al.(2020)Huang, Hu, Yeung, and Chen]{huang2020conv}
Wenyong Huang, Wenchao Hu, Yu~Ting Yeung, and Xiao Chen.
\newblock Conv-transformer transducer: Low latency, low frame rate, streamable
  end-to-end speech recognition.
\newblock \emph{arXiv preprint arXiv:2008.05750}, 2020.

\bibitem[Ioffe \& Szegedy(2015)Ioffe and Szegedy]{ioffe2015batch}
Sergey Ioffe and Christian Szegedy.
\newblock Batch normalization: Accelerating deep network training by reducing
  internal covariate shift.
\newblock \emph{arXiv preprint arXiv:1502.03167}, 2015.

\bibitem[Jaitly et~al.(2016)Jaitly, Le, Vinyals, Sutskever, Sussillo, and
  Bengio]{jaitly2016neural}
Navdeep Jaitly, Quoc~V Le, Oriol Vinyals, Ilya Sutskever, David Sussillo, and
  Samy Bengio.
\newblock An online sequence-to-sequence model using partial conditioning.
\newblock In \emph{Advances in Neural Information Processing Systems 29}, pp.\
  5067--5075, 2016.

\bibitem[Kannan et~al.(2019)Kannan, Datta, Sainath, Weinstein, Ramabhadran, Wu,
  Bapna, Chen, and Lee]{kannan2019large}
Anjuli Kannan, Arindrima Datta, Tara~N Sainath, Eugene Weinstein, Bhuvana
  Ramabhadran, Yonghui Wu, Ankur Bapna, Zhifeng Chen, and Seungji Lee.
\newblock Large-scale multilingual speech recognition with a streaming
  end-to-end model.
\newblock \emph{arXiv preprint arXiv:1909.05330}, 2019.

\bibitem[Kingma \& Ba(2014)Kingma and Ba]{kingma2014adam}
Diederik~P Kingma and Jimmy Ba.
\newblock Adam: A method for stochastic optimization.
\newblock \emph{arXiv preprint arXiv:1412.6980}, 2014.

\bibitem[Li et~al.(2020{\natexlab{a}})Li, Chang, Sainath, Pang, He, Strohman,
  and Wu]{li2020towards}
Bo~Li, Shuo-yiin Chang, Tara~N Sainath, Ruoming Pang, Yanzhang He, Trevor
  Strohman, and Yonghui Wu.
\newblock Towards fast and accurate streaming end-to-end asr.
\newblock In \emph{ICASSP 2020-2020 IEEE International Conference on Acoustics,
  Speech and Signal Processing (ICASSP)}, pp.\  6069--6073. IEEE,
  2020{\natexlab{a}}.

\bibitem[Li et~al.(2020{\natexlab{b}})Li, Qin, Chiu, Pang, and
  He]{li2020parallel}
Wei Li, James Qin, Chung-Cheng Chiu, Ruoming Pang, and Yanzhang He.
\newblock Parallel rescoring with transformer for streaming on-device speech
  recognition.
\newblock \emph{arXiv preprint arXiv:2008.13093}, 2020{\natexlab{b}}.

\bibitem[Miao et~al.(2020)Miao, Cheng, Gao, Zhang, and
  Yan]{miao2020transformer}
Haoran Miao, Gaofeng Cheng, Changfeng Gao, Pengyuan Zhang, and Yonghong Yan.
\newblock Transformer-based online ctc/attention end-to-end speech recognition
  architecture.
\newblock In \emph{ICASSP 2020-2020 IEEE International Conference on Acoustics,
  Speech and Signal Processing (ICASSP)}, pp.\  6084--6088. IEEE, 2020.

\bibitem[Moritz et~al.(2019)Moritz, Hori, and Le~Roux]{moritz2019triggered}
Niko Moritz, Takaaki Hori, and Jonathan Le~Roux.
\newblock Triggered attention for end-to-end speech recognition.
\newblock In \emph{ICASSP 2019-2019 IEEE International Conference on Acoustics,
  Speech and Signal Processing (ICASSP)}, pp.\  5666--5670. IEEE, 2019.

\bibitem[Moritz et~al.(2020)Moritz, Hori, and Le]{moritz2020streaming}
Niko Moritz, Takaaki Hori, and Jonathan Le.
\newblock Streaming automatic speech recognition with the transformer model.
\newblock In \emph{ICASSP 2020-2020 IEEE International Conference on Acoustics,
  Speech and Signal Processing (ICASSP)}, pp.\  6074--6078. IEEE, 2020.

\bibitem[Narayanan et~al.(2018)Narayanan, Misra, Sim, Pundak, Tripathi, Elfeky,
  Haghani, Strohman, and Bacchiani]{narayanan2018toward}
Arun Narayanan, Ananya Misra, Khe~Chai Sim, Golan Pundak, Anshuman Tripathi,
  Mohamed Elfeky, Parisa Haghani, Trevor Strohman, and Michiel Bacchiani.
\newblock Toward domain-invariant speech recognition via large scale training.
\newblock In \emph{2018 IEEE Spoken Language Technology Workshop (SLT)}, pp.\
  441--447. IEEE, 2018.

\bibitem[Narayanan et~al.(2020)Narayanan, Sainath, Pang, Yu, Chiu,
  Prabhavalkar, Variani, and Strohman]{narayanan2020cascaded}
Arun Narayanan, Tara~N Sainath, Ruoming Pang, Jiahui Yu, Chung-Cheng Chiu,
  Rohit Prabhavalkar, Ehsan Variani, and Trevor Strohman.
\newblock Cascaded encoders for unifying streaming and non-streaming asr.
\newblock \emph{arXiv preprint arXiv:2010.14606}, 2020.

\bibitem[Oord et~al.(2016)Oord, Dieleman, Zen, Simonyan, Vinyals, Graves,
  Kalchbrenner, Senior, and Kavukcuoglu]{oord2016wavenet}
Aaron van~den Oord, Sander Dieleman, Heiga Zen, Karen Simonyan, Oriol Vinyals,
  Alex Graves, Nal Kalchbrenner, Andrew Senior, and Koray Kavukcuoglu.
\newblock Wavenet: A generative model for raw audio.
\newblock \emph{arXiv preprint arXiv:1609.03499}, 2016.

\bibitem[Panayotov et~al.(2015)Panayotov, Chen, Povey, and
  Khudanpur]{panayotov2015librispeech}
Vassil Panayotov, Guoguo Chen, Daniel Povey, and Sanjeev Khudanpur.
\newblock Librispeech: an asr corpus based on public domain audio books.
\newblock In \emph{2015 IEEE International Conference on Acoustics, Speech and
  Signal Processing (ICASSP)}, pp.\  5206--5210. IEEE, 2015.

\bibitem[Panchapagesan et~al.(2020)Panchapagesan, Park, Chiu, Shangguan, Liang,
  and Gruenstein]{panchapagesan2020efficient}
Sankaran Panchapagesan, Daniel~S Park, Chung-Cheng Chiu, Yuan Shangguan, Qiao
  Liang, and Alexander Gruenstein.
\newblock Efficient knowledge distillation for rnn-transducer models.
\newblock \emph{arXiv preprint arXiv:2011.06110}, 2020.

\bibitem[Papernot et~al.(2016)Papernot, McDaniel, Wu, Jha, and
  Swami]{papernot2016distillation}
Nicolas Papernot, Patrick McDaniel, Xi~Wu, Somesh Jha, and Ananthram Swami.
\newblock Distillation as a defense to adversarial perturbations against deep
  neural networks.
\newblock In \emph{2016 IEEE Symposium on Security and Privacy (SP)}, pp.\
  582--597. IEEE, 2016.

\bibitem[Park et~al.(2019)Park, Chan, Zhang, Chiu, Zoph, Cubuk, and
  Le]{park2019specaugment}
Daniel~S Park, William Chan, Yu~Zhang, Chung-Cheng Chiu, Barret Zoph, Ekin~D
  Cubuk, and Quoc~V Le.
\newblock Specaugment: A simple data augmentation method for automatic speech
  recognition.
\newblock \emph{arXiv preprint arXiv:1904.08779}, 2019.

\bibitem[Prabhavalkar et~al.(2018)Prabhavalkar, Sainath, Wu, Nguyen, Chen,
  Chiu, and Kannan]{prabhavalkar2018minimum}
Rohit Prabhavalkar, Tara~N Sainath, Yonghui Wu, Patrick Nguyen, Zhifeng Chen,
  Chung-Cheng Chiu, and Anjuli Kannan.
\newblock Minimum word error rate training for attention-based
  sequence-to-sequence models.
\newblock In \emph{2018 IEEE International Conference on Acoustics, Speech and
  Signal Processing (ICASSP)}, pp.\  4839--4843. IEEE, 2018.

\bibitem[Raffel et~al.(2017)Raffel, Luong, Liu, Weiss, and Eck]{colin17}
C.~Raffel, M.~Luong, P.~J. Liu, R.J. Weiss, and D.~Eck.
\newblock {Online and Linear-Time Attention by Enforcing Monotonic Alignments}.
\newblock In \emph{Proc. ICML}, 2017.

\bibitem[Ramachandran et~al.(2017)Ramachandran, Zoph, and
  Le]{ramachandran2017searching}
Prajit Ramachandran, Barret Zoph, and Quoc~V Le.
\newblock Searching for activation functions.
\newblock \emph{arXiv preprint arXiv:1710.05941}, 2017.

\bibitem[Sainath et~al.(2019)Sainath, Pang, Rybach, He, Prabhavalkar, Li,
  Visontai, Liang, Strohman, Wu, et~al.]{sainath2019two}
Tara~N Sainath, Ruoming Pang, David Rybach, Yanzhang He, Rohit Prabhavalkar,
  Wei Li, Mirk{\'o} Visontai, Qiao Liang, Trevor Strohman, Yonghui Wu, et~al.
\newblock Two-pass end-to-end speech recognition.
\newblock \emph{arXiv preprint arXiv:1908.10992}, 2019.

\bibitem[Sainath et~al.(2020)Sainath, He, Li, Narayanan, Pang, Bruguier, Chang,
  Li, Alvarez, Chen, et~al.]{sainath2020streaming}
Tara~N Sainath, Yanzhang He, Bo~Li, Arun Narayanan, Ruoming Pang, Antoine
  Bruguier, Shuo-yiin Chang, Wei Li, Raziel Alvarez, Zhifeng Chen, et~al.
\newblock A streaming on-device end-to-end model surpassing server-side
  conventional model quality and latency.
\newblock In \emph{ICASSP 2020-2020 IEEE International Conference on Acoustics,
  Speech and Signal Processing (ICASSP)}, pp.\  6059--6063. IEEE, 2020.

\bibitem[Shen et~al.(2019)Shen, Nguyen, Wu, Chen, Chen, Jia, Kannan, Sainath,
  Cao, Chiu, et~al.]{shen2019lingvo}
Jonathan Shen, Patrick Nguyen, Yonghui Wu, Zhifeng Chen, Mia~X Chen, Ye~Jia,
  Anjuli Kannan, Tara Sainath, Yuan Cao, Chung-Cheng Chiu, et~al.
\newblock Lingvo: a modular and scalable framework for sequence-to-sequence
  modeling.
\newblock \emph{arXiv preprint arXiv:1902.08295}, 2019.

\bibitem[Srivastava et~al.(2014)Srivastava, Hinton, Krizhevsky, Sutskever, and
  Salakhutdinov]{srivastava2014dropout}
Nitish Srivastava, Geoffrey Hinton, Alex Krizhevsky, Ilya Sutskever, and Ruslan
  Salakhutdinov.
\newblock Dropout: a simple way to prevent neural networks from overfitting.
\newblock \emph{The journal of machine learning research}, 15\penalty0
  (1):\penalty0 1929--1958, 2014.

\bibitem[Tsunoo et~al.(2019)Tsunoo, Kashiwagi, Kumakura, and
  Watanabe]{tsunoo2019towards}
Emiru Tsunoo, Yosuke Kashiwagi, Toshiyuki Kumakura, and Shinji Watanabe.
\newblock Towards online end-to-end transformer automatic speech recognition.
\newblock \emph{arXiv preprint arXiv:1910.11871}, 2019.

\bibitem[Tsunoo et~al.(2020)Tsunoo, Kashiwagi, and
  Watanabe]{tsunoo2020streaming}
Emiru Tsunoo, Yosuke Kashiwagi, and Shinji Watanabe.
\newblock Streaming transformer asr with blockwise synchronous inference.
\newblock \emph{arXiv preprint arXiv:2006.14941}, 2020.

\bibitem[Tzeng et~al.(2015)Tzeng, Hoffman, Darrell, and
  Saenko]{tzeng2015simultaneous}
Eric Tzeng, Judy Hoffman, Trevor Darrell, and Kate Saenko.
\newblock Simultaneous deep transfer across domains and tasks.
\newblock In \emph{Proceedings of the IEEE International Conference on Computer
  Vision}, pp.\  4068--4076, 2015.

\bibitem[Vaswani et~al.(2017)Vaswani, Shazeer, Parmar, Uszkoreit, Jones, Gomez,
  Kaiser, and Polosukhin]{vaswani2017attention}
Ashish Vaswani, Noam Shazeer, Niki Parmar, Jakob Uszkoreit, Llion Jones,
  Aidan~N Gomez, {\L}ukasz Kaiser, and Illia Polosukhin.
\newblock Attention is all you need.
\newblock In \emph{Advances in neural information processing systems}, pp.\
  5998--6008, 2017.

\bibitem[Wang et~al.(2020)Wang, Wu, Liu, Li, Lu, Ye, and Zhou]{wang2020low}
Chengyi Wang, Yu~Wu, Shujie Liu, Jinyu Li, Liang Lu, Guoli Ye, and Ming Zhou.
\newblock Low latency end-to-end streaming speech recognition with a scout
  network.
\newblock \emph{arXiv preprint arXiv:2003.10369}, 2020.

\bibitem[Watanabe et~al.(2017)Watanabe, Hori, Kim, Hershey, and
  Hayashi]{watanabe2017hybrid}
Shinji Watanabe, Takaaki Hori, Suyoun Kim, John~R Hershey, and Tomoki Hayashi.
\newblock Hybrid ctc/attention architecture for end-to-end speech recognition.
\newblock \emph{IEEE Journal of Selected Topics in Signal Processing},
  11\penalty0 (8):\penalty0 1240--1253, 2017.

\bibitem[Williams \& Zipser(1989)Williams and Zipser]{williams1989learning}
Ronald~J Williams and David Zipser.
\newblock A learning algorithm for continually running fully recurrent neural
  networks.
\newblock \emph{Neural computation}, 1\penalty0 (2):\penalty0 270--280, 1989.

\bibitem[Wu et~al.(2016)Wu, Schuster, Chen, Le, Norouzi, Macherey, Krikun, Cao,
  Gao, Macherey, et~al.]{wu2016google}
Yonghui Wu, Mike Schuster, Zhifeng Chen, Quoc~V Le, Mohammad Norouzi, Wolfgang
  Macherey, Maxim Krikun, Yuan Cao, Qin Gao, Klaus Macherey, et~al.
\newblock Google's neural machine translation system: Bridging the gap between
  human and machine translation.
\newblock \emph{arXiv preprint arXiv:1609.08144}, 2016.

\bibitem[Wu et~al.(2020)Wu, Zhao, Liang, Yu, Gulati, and Pang]{wu2020dynamic}
Zhaofeng Wu, Ding Zhao, Qiao Liang, Jiahui Yu, Anmol Gulati, and Ruoming Pang.
\newblock Dynamic sparsity neural networks for automatic speech recognition.
\newblock \emph{arXiv preprint arXiv:2005.10627}, 2020.

\bibitem[Yeh et~al.(2019)Yeh, Mahadeokar, Kalgaonkar, Wang, Le, Jain, Schubert,
  Fuegen, and Seltzer]{yeh2019transformer}
Ching-Feng Yeh, Jay Mahadeokar, Kaustubh Kalgaonkar, Yongqiang Wang, Duc Le,
  Mahaveer Jain, Kjell Schubert, Christian Fuegen, and Michael~L Seltzer.
\newblock Transformer-transducer: End-to-end speech recognition with
  self-attention.
\newblock \emph{arXiv preprint arXiv:1910.12977}, 2019.

\bibitem[Yu \& Huang(2019{\natexlab{a}})Yu and Huang]{yu2019autoslim}
Jiahui Yu and Thomas Huang.
\newblock Autoslim: Towards one-shot architecture search for channel numbers.
\newblock \emph{arXiv preprint arXiv:1903.11728}, 2019{\natexlab{a}}.

\bibitem[Yu \& Huang(2019{\natexlab{b}})Yu and Huang]{yu2019universally}
Jiahui Yu and Thomas~S Huang.
\newblock Universally slimmable networks and improved training techniques.
\newblock In \emph{Proceedings of the IEEE International Conference on Computer
  Vision}, pp.\  1803--1811, 2019{\natexlab{b}}.

\bibitem[Yu et~al.(2018)Yu, Yang, Xu, Yang, and Huang]{yu2018slimmable}
Jiahui Yu, Linjie Yang, Ning Xu, Jianchao Yang, and Thomas Huang.
\newblock Slimmable neural networks.
\newblock In \emph{International Conference on Learning Representations}, 2018.

\bibitem[Yu et~al.(2020)Yu, Jin, Liu, Bender, Kindermans, Tan, Huang, Song,
  Pang, and Le]{yu2020bignas}
Jiahui Yu, Pengchong Jin, Hanxiao Liu, Gabriel Bender, Pieter-Jan Kindermans,
  Mingxing Tan, Thomas Huang, Xiaodan Song, Ruoming Pang, and Quoc Le.
\newblock Bignas: Scaling up neural architecture search with big single-stage
  models.
\newblock \emph{arXiv preprint arXiv:2003.11142}, 2020.

\bibitem[Zhang et~al.(2020)Zhang, Lu, Sak, Tripathi, McDermott, Koo, and
  Kumar]{zhang2020transformer}
Qian Zhang, Han Lu, Hasim Sak, Anshuman Tripathi, Erik McDermott, Stephen Koo,
  and Shankar Kumar.
\newblock Transformer transducer: A streamable speech recognition model with
  transformer encoders and rnn-t loss.
\newblock In \emph{ICASSP 2020-2020 IEEE International Conference on Acoustics,
  Speech and Signal Processing (ICASSP)}, pp.\  7829--7833. IEEE, 2020.

\end{thebibliography}
